\documentclass[10pt,twocolumn,letterpaper]{article}
\usepackage{multirow} 
\usepackage{soul}
\usepackage{pifont}
\usepackage{stfloats}
\usepackage{float}
\usepackage{algorithmic}
\usepackage{algorithm}
\usepackage{hhline}
\usepackage{setspace}
\usepackage[table]{xcolor}
\usepackage{cvpr}              
\definecolor{cvprblue}{rgb}{0.21,0.49,0.74}
\usepackage[pagebackref,breaklinks,colorlinks,allcolors=cvprblue]{hyperref}

\def\paperID{xxxx} 
\def\confName{CVPR}
\def\confYear{2026}

\title{BarbieGait: An Identity-Consistent Synthetic Human Dataset with \\
Versatile Cloth-Changing for Gait Recognition}

\author{
    Qingyuan Cai$^{1}$ \hfill
    Saihui Hou$^{1}$ \hfill
    Xuecai Hu$^{2}$ \hfill
    Yongzhen Huang$^{1,3}$\textsuperscript{*}   \\
    $^{1}$School of Artificial Intelligence, Beijing Normal University \\
    $^{2}$AMAP, Alibaba Group  \hspace{1.3em}  $^{3}$WATRIX.AI \\
    {\tt\small caiqingyuan@mail.bnu.edu.cn,
    huxc@mail.ustc.edu.cn,
    \{housaihui, huangyongzhen\}@bnu.edu.cn }
}

\begin{document}

\twocolumn[{%
\renewcommand\twocolumn[1][]{#1}%
\maketitle 
\begin{center}
    \centering
    \vspace{-10mm}
    \includegraphics[width=\linewidth]{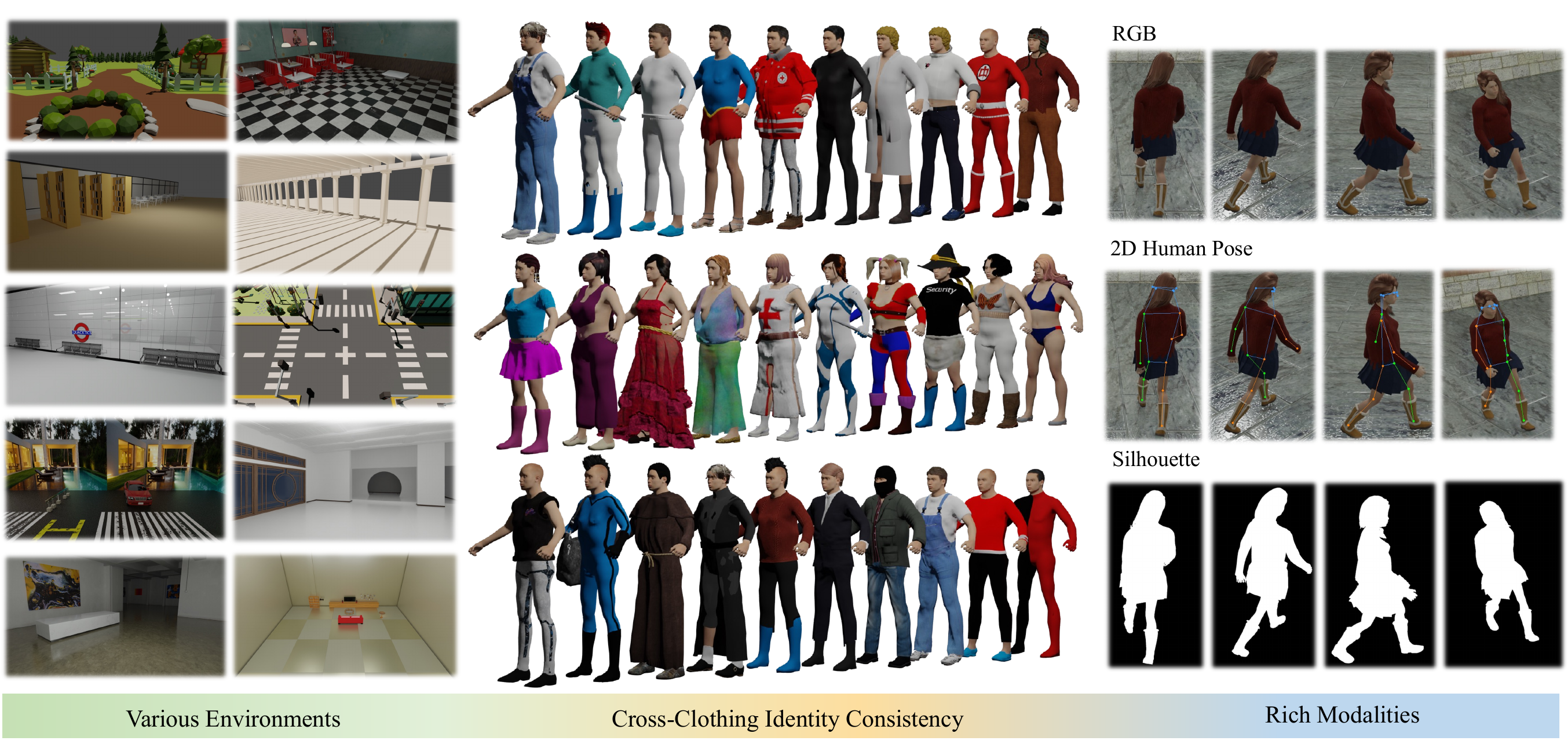}
    \setlength{\abovecaptionskip}{0mm}
    \vspace{-0.4cm} 
    \captionof{figure}{
    BarbieGait is an identity-consistent synthetic human dataset, where each subject has 100 different kinds of clothes combinations, including variations in hairstyles, upper and lower garments, and carried accessories. We simulate diverse illumination in various environments to generate RGB images and extract multimodal data (2D Human Pose, Silhouette) for gait recognition.
	}
    \label{fig1}
\end{center}
}]

\begingroup\renewcommand{\thefootnote}{\fnsymbol{footnote}}\footnotetext[1]{Corresponding author.}\endgroup
\begin{abstract}
    Gait recognition, as a reliable biometric technology, has seen rapid development in recent years while it faces significant challenges caused by diverse clothing styles in the real world. This paper introduces \textbf{BarbieGait}, a synthetic gait dataset where real-world subjects are uniquely mapped into a virtual engine to simulate extensive clothing changes while preserving their gait identity information. As a pioneering work, BarbieGait provides a controllable gait data generation method, enabling the production of large datasets to validate cross-clothing issues that are difficult to verify with real-world data. However, the diversity of clothing increases intra-class variance and makes one of the biggest challenges to learning cloth-invariant features under varying clothing conditions. Therefore, we propose \textbf{GaitCLIF} (Gait-oriented CLoth-Invariant Feature) as a robust baseline model for cross-clothing gait recognition. Through extensive experiments, we validate that our method significantly improves cross-clothing performance on BarbieGait and the existing popular gait benchmarks. We believe that BarbieGait, with its extensive cross-clothing gait data, will further advance the capabilities of gait recognition in cross-clothing scenarios and promote progress in related research. The source code and dataset are available at \url{https://github.com/BarbieGait/BarbieGait}.
\end{abstract}    
\section{Introduction}
\label{sec:intro}
Gait recognition is a reliable biometric technology that identifies subjects based on their walking patterns. It can be performed from a distance without requiring direct interaction with the subject, making it ideal for surveillance and security applications. In recent works, the research on gait recognition has grown fast. However, human appearance variations caused by covariates such as clothing and carrying \cite{li2023depth,shen2023lidargait} remain a major bottleneck in the development of gait recognition. Although many benchmarks are proposed~\cite{yu2006framework,yu2006framework,takemura2018multi,zheng2022gait,zhu2021gait,wang2025ra,meng2025seeing,dong2024hybridgait}, such as some in-the-lab datasets CASIA-B \cite{yu2006framework}, OU-MVLP \cite{takemura2018multi}, and in-the-wild dataset Gait3D \cite{zheng2022gait}, GREW \cite{zhu2021gait} et al., they still lack extensive cloth-changing data for every subject. The newly proposed cross-clothing dataset CCPG \cite{li2023depth}, investing efforts and costs in collecting cross-clothing data for each subject, only has seven different cloth-changing statuses per person. Without a large amount of cloth-changing data, it becomes difficult to prove whether gait recognition is reliable under extensive clothing variations or if there is still room for improvement. Moreover, the limited clothing diversity makes it insufficient to demonstrate whether existing methods can effectively handle scenarios involving extensive clothing changes. However, collecting cross-clothing gait data that encompasses diverse and complex clothing styles across various ethnicities and seasons is not only immensely costly but also nearly impossible to achieve due to privacy concerns.
\par Nowadays, some human-centric tasks like Human Pose and Shape Estimation~\cite{bazavan2021hspace, cai2024playing, patel2021agora, cai2024disentangled, cai2025fastddhpose}, Human Neural Rendering~\cite{yang2023synbody}, Person Re-Identification~\cite{wang2020surpassing, zhang2021unrealperson}, and Gait Recognition \cite{zhang2023large} attempt to generate frame-level synthetic images to reduce data collection costs and get massive data. While these synthetic datasets encompass a wide range of motion sequences, they primarily emphasize action diversity, overlooking a fundamental requirement for gait recognition research. The key question we seek to address is: \textit{\textbf{can these generative paradigms be used to synthesize cloth-changing gait data while preserving the gait identity discriminability of a real subject?}}
\par Motivated by the key question, we develop a new synthetic gait dataset named \textbf{BarbieGait} for cloth-changing gait recognition research which has two main characteristics: \textit{\textbf{(1) Each virtual character is authentically replicated from a real subject through dual subject-specific alignment protocols: skeleton length and body shape matching, combined with kinematic motion matching. (2) Each subject has 100 completely random full-body clothing changes, yielding highly diverse clothing variations.}}  To the best of our knowledge, we are the first to use high-precision 3D human pose and mesh to achieve real subjects and virtual subjects alignment in both static and dynamic levels, preserving unique gait identity in the generated cloth-changing gait sequence from the real subject. This alignment approach prevents the problem in prior methods \cite{bazavan2021hspace, yang2023synbody, cai2024playing, zhang2023large}, where synthesized gait sequences of the same individual fail to maintain identity consistency due to the reuse of identical motions across different subjects or the use of diverse gait patterns for the same subject.


\par BarbieGait, as a versatile dataset, provides a solid foundation for addressing challenges of cloth-changing and opens up new possibilities for exploring various aspects of gait recognition, such as cross-view and cross-domain issues in the future due to the controllable data generation paradigm. In this paper, we not only utilize BarbieGait as a comprehensive benchmark but also focus on learning \emph{cloth-invariant} features under cloth-changing conditions. The diversity of cloth-changing significantly increases the intra-class variance for the same identity and hinders the model's ability to extract identity-related features from cross-clothing features. Therefore, we believe the primary issue is to eliminate cloth-specific statistics. Another key perspective is that, based on human dressing characteristics, features from different parts of the body are affected by clothing to varying degrees. Preserving fine-grained motion information will be more conducive to learning cloth-invariant features from different parts of the human body. Based on these insights, we propose a straightforward yet powerful approach, \textbf{GaitCLIF} (Gait-oriented CLoth-Invariant Feature), which serves as a robust baseline model for gait recognition in cloth-changing conditions. Furthermore, GaitCLIF demonstrates consistent performance improvements on BarbieGait and existing gait recognition benchmarks.

\par Our main contributions can be summarized as follows:
\begin{itemize}
    \item We present \textbf{BarbieGait}, a synthetic gait recognition dataset containing 521 subjects, each with 100 clothing variations generated under consistent identity. It sets a new benchmark with greater clothing diversity and more gait sequences than existing datasets.
\end{itemize}

\begin{itemize}
    \item To tackle the increased intra-class variance caused by diverse clothing changes, we propose \textbf{GaitCLIF}, a robust baseline for learning fine-grained \emph{cloth-invariant} motion features for cloth-changing gait recognition.
    \end{itemize}

\begin{itemize}
    \item We demonstrate that our method can notably improve performance on BarbieGait and achieve state-of-the-art performance on the existing datasets, including CCPG, SUSTech1K, Gait3D, and GREW, which involve clothing variations and style changes.
\end{itemize}
\section{Related Work}
\subsection{Human Synthetic Dataset}
Existing synthetic datasets mainly serve Human Pose Estimation, Human Mesh Recovery, and Person Re-Identification. SURREAL~\cite{varol2017learning} renders clothing onto SMPL~\cite{loper2023smpl} bodies but lacks accurate clothing-body alignment. AGORA~\cite{patel2021agora} provides high-quality real scans but lacks continuous motion due to the high cost of scanning. SynBody~\cite{yang2023synbody} and GTA-Human~\cite{cai2024playing} animates parametric models (e.g., GHUM~\cite{xu2020ghum}, SMPL, SMPL-X~\cite{pavlakos2019expressive}) to generate video sequences. RandPerson~\cite{wang2020surpassing}, UnrealPerson~\cite{zhang2021unrealperson}, and SynPerson~\cite{xiang2022rethinking} focus on synthetic Re-ID training. VersatileGait~\cite{zhang2023large} targets gait recognition, showing the potential of synthetic data. However, existing datasets often fail to preserve gait identity due to two common issues: (1) the same motion sequence is used to drive different subjects, and (2) the same subject is animated using motion sequences with significant variation in walking patterns.

\begin{figure*}[t] 
    \centering
    \includegraphics[width=1.0\textwidth]{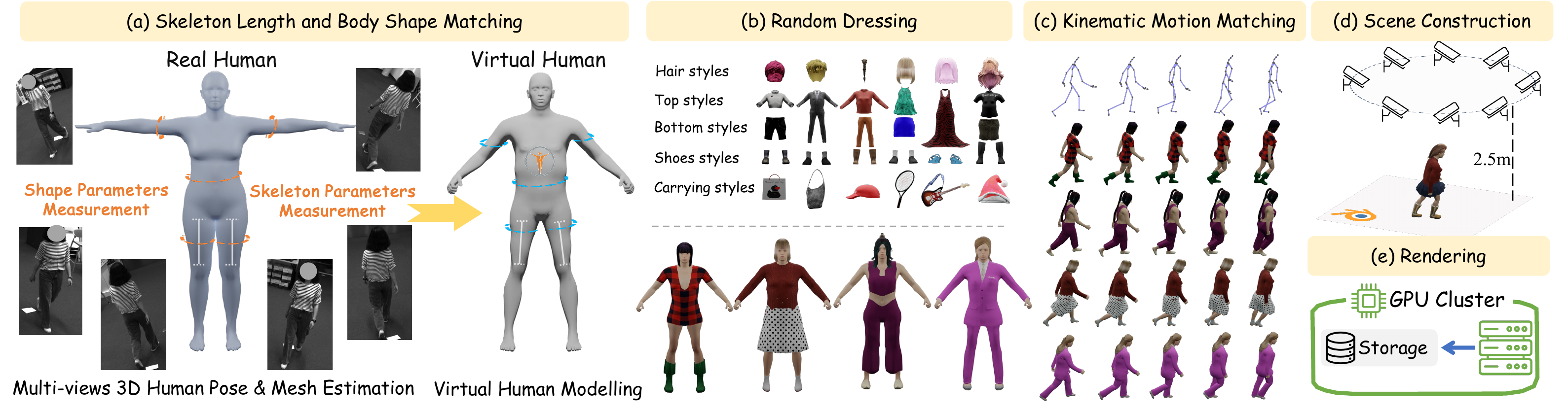}  
    \vspace{-0.7cm} 
    \caption{The BarbieGait data generation system includes:  (a) \emph{Skeleton Length and Body Shape Matching} maps real humans to virtual ones based on 3D skeleton and body shape using MakeHuman \cite{makehuman}. (b) \emph{Random Dressing} refers to randomly selecting outfits for cloth-changing. (c) \emph{Kinematic Motion Matching} refers to the alignment of gait identity information across different outfits of the same real subject. (d) \emph{Scene Construction} uses Blender \cite{blender} to create multiple environments and capture multi-view images. (e) \emph{Rendering} refers to accelerate image generation using a GPU cluster. }  
    \label{fig2}  
    \vspace{-0.4cm}
\end{figure*}

\subsection{Gait Recognition}
The existing gait recognition methods can be categorized into appearance-based methods and pose-based methods.

\par\noindent \textbf{Appearance-based methods}: Appearance-based methods primarily rely on human silhouettes or RGB visual appearance cues for gait recognition. Representative silhouette-based methods~\cite{chao2019gaitset,fan2020gaitpart,lin2022gaitgl,fan2023opengait,fan2025opengait, fan2023exploring,li2024aerialgait,li2023gait,huang2025vocabulary,huang2025learning,huang2024occluded,yang2025bridging,hou2025gaitsnippet} mainly focus on learning spatio-temporal dynamic patterns from silhouette sequences. In addition, RGB-based methods, including BigGait~\cite{ye2024biggait}, BiggerGait~\cite{ye2025biggergait}, DenoisingGait~\cite{jin2025denoising}, and Gait-X~\cite{wang2025gait}, exploit informative appearance and motion cues from RGB inputs to compensate for the limitations of using silhouettes alone. However, regardless of whether silhouette or RGB modalities are used, clothing changes introduce substantial appearance variations and significantly increase the difficulty of gait recognition.


\par\noindent \textbf{Pose-based methods}: 
Pose-based methods are initially proposed to emphasize motion patterns in gait recognition~\cite{liao2020model,teepe2021gaitgraph,teepe2022towards,zhang2023spatial,lang2025beyond}. However, the sparse nature of human keypoints often leads to overfitting and poor generalization. GPGait~\cite{fu2023gpgait} addresses this issue by proposing a more robust framework to enhance the generalization ability of pose-based methods, achieving promising results. To obtain denser pose information, SkeletonGait~\cite{fan2024skeletongait} and GaitHeat~\cite{fu2025cut} use heatmaps, while DPGait~\cite{lang2025beyond} uses dense keypoints to achieve a more comprehensive representation of human shape.


\section{BarbieGait}
BarbieGait is designed to provide rich cloth-changing gait data to improve the capability of cross-clothing gait recognition. The dataset is built on two principles: (1) \textbf{Uniquely Identity Mapping}: It ensures each generated cloth-changing gait sequence preserves the unique identity of a real subject. (2) \textbf{Cloth-Changing Quantity}: A large volume of clothing variations is essential for improving and validating recognition capabilities. 

\begin{table*}[ht]
    \centering
    \renewcommand{\arraystretch}{1} 
    \caption{Comparison with existing Motion, Person Re-ID, and Gait datasets. “\#Subj’’ denotes the number of subjects corresponding to real identities. “\#Mesh”, “\#Views”, and “\#Seq’’ represent the total number of meshes, camera viewpoints, and sequences, respectively. “Cloth-Changing’’ indicates whether subjects appear in multiple outfits, and “\#Cloth/Subj.’’ specifies the number of clothing variations per subject. “GT Format’’ lists available ground-truth types: “3DJ’’ (3D joints), “3DM’’ (3D meshes), and “silh.’’ (silhouettes).}

    \resizebox{\textwidth}{!}{
    \begin{tabular}{clccccccccc}
    
    \toprule
    \multicolumn{1}{l}{}    & \textbf{Dataset}      & \textbf{Year} & \textbf{Type}      & \textbf{\#Subj}.  & \textbf{\#Mesh}         & \textbf{\#Views} & \textbf{\#Seq}     & \textbf{Cloth-Changing} & \textbf{\#Cloth/Subj.} & \textbf{GT Format}       \\ \midrule
    \multirow{4}{*}{Motion} 
                            & SURREAL \cite{varol2017learning}      & 2017 & Synthetic & \textemdash  & 145              & 1       & NA        & $\times$   & \textemdash                     & 3DM             \\
                            & AGORA \cite{patel2021agora}        & 2021 & Synthetic & \textemdash  & \textgreater{}350 & 1       & NA        & $\times$      & \textemdash                  & 3DM, silh.      \\
                            & GTA-Human \cite{cai2024playing}    & 2021 & Synthetic & \textemdash  & \textgreater{}600 & 1       & 20K       & $\times$      & \textemdash                 & 3DM             \\
                            & SynBody \cite{yang2023synbody}      & 2023 & Synthetic & \textemdash & 10,000            & 4       & 40K       & $\times$     & \textemdash                   & 3DM, silh.      \\ \midrule
    \multirow{3}{*}{Re-ID}  & RandPerson \cite{wang2020surpassing}    & 2020 & Synthetic & \textemdash  & 8,000             & 19      & NA     & $\times$    & \textemdash         & NA              \\
                            & UnrealPerson \cite{zhang2021unrealperson}  & 2021 & Synthetic & \textemdash  & 3,000          & 34      & NA     & $\times$    & \textemdash         & NA              \\
                            & SynPerson \cite{xiang2022rethinking}     & 2022 & Synthetic & \textemdash  & 5,345            & 36      & NA     & $\times$    & \textemdash         & NA              \\ \midrule
    \multirow{7}{*}{Gait}   & CASIA-B \cite{yu2006framework}      & 2006 & Real      & 124  & \textemdash             & 11      & 13,640    & $\surd$     & 3                   & NA              \\
                            & OU-MVLP \cite{takemura2018multi}      & 2018 & Real      & 10,307  & \textemdash          & 14      & 288,596   & $\times$   & \textemdash                     & NA              \\
                            & GREW \cite{zhu2021gait}         & 2021 & Real      & 26,345   & \textemdash         & 882     & 128,671   & $\surd$         & 6               & NA              \\
                            & Gait3D \cite{zheng2022gait}       & 2022 & Real      & 4,000  & \textemdash           & 39      & 25,309    & $\times$       & \textemdash                 & NA              \\
                            & CASIA-E \cite{song2022casia}      & 2023 & Real      & 1,014  & \textemdash            & 26      & 778,752   & $\surd$      & 3                  & NA              \\
                            & VersatileGait \cite{zhang2023large} & 2023 & Synthetic & \textemdash  & 10,000            & 44      & 1,320,000 & $\times$     & \textemdash           & NA             \\
                            & CCPG \cite{li2023depth}         & 2023 & Real      & 200  & \textemdash             & 10      & 16,566    & $\surd$         & 7               & NA              \\
                            \rowcolor{blue!10}
                            \cellcolor{white}& \textbf{BarbieGait (Ours)}      & \textbf{2025} & \textbf{Synthetic} & \textbf{521}  & \textbf{521,000}      & \textbf{8}       & \textbf{1,203,324} & $\surd$                & \textbf{100}      & \textbf{3DJ, 3DM, silh.} \\ \bottomrule
    
    \end{tabular}
    }
    
    \label{tab1}
    \vspace{-0.3cm}
    \end{table*}

\subsection{Cornerstone: Raw Data Collection}
As a synthetic gait dataset with various cloth-changing scenarios, one of the key advantages of BarbieGait is that it ensures the gait identity information of each virtual character comes from a real human. 
\par We deploy a synchronized camera array comprising six cameras and collect a real-world gait dataset of 521 subjects, with three gait sequences recorded for each subject, as shown in Figure \ref{fig2}a. Subsequently, we first estimate 2D poses from multi-views using our pretrained HRNet~\cite{wang2020deep}. Then, for each gait sequence, we estimate 3D human pose and mesh through triangulation \cite{qiu2019cross} and EasyMoCap \cite{easymocap}. High-quality 3D human pose and mesh serve as the cornerstone for ensuring the unique identity information is replicated from a real subject to virtual subjects.


\subsection{BarbieGait Generation System}
In this subsection, we introduce our data synthesis system in Figure \ref{fig2} for generating the BarbieGait dataset based on the collected real-world data. The system consists of five components: (1) a skeleton length and body shape matching module to generate a virtual human aligned with the real human, (2) a random dressing module to change 100 outfits for each subject, (3) a kinematic motion matching module to preserve the walking patterns consistency between virtual human with various cloth-changing and the real human. (4) a scene construction module to simulate real-world collection conditions, and (5) a rendering module.

\vspace{-0.3cm}
\paragraph{Skeleton Length and Body Shape Matching.}
Skeleton Length and Body Shape Matching shown in Figure \ref{fig2}(a) refers to the alignment between the real-world subject and the MakeHuman \cite{makehuman} generated virtual human. It consists of two aspects: (1) Skeleton Length: Since 3D skeleton length is basically stable for a subject, we align the virtual human's skeleton length, such as thigh and lower leg lengths, with the real skeleton based on 3D poses. (2) Body Shape: Using EasyMoCap \cite{easymocap}, we estimate SMPL \cite{loper2023smpl} mesh for each frame. To minimize errors in human mesh recovery, we define 12 static circumference parameters (e.g., neck, chest, waist, hip, etc.) and use their frame-averaged values to align each subject's virtual body.

\vspace{-0.5cm}

\paragraph{Random Dressing.}
Random Dressing in Figure \ref{fig2}b enables diverse clothing changes in a controlled manner. We use MakeHuman to randomly select 100 outfits for each subject from a diverse wardrobe, following seasonal and practical daily clothing matching strategies.

\vspace{-0.3cm}
\paragraph{Kinematic Motion Matching.}


Kinematic motion matching is a widely adopted technique in character animation systems, where rig-to-rig motion transfer is achieved by mapping local bone transformations and applying them frame-by-frame to the target armature~\cite{blender_animation_retargeting,Rokoko}. 

Building on these principles, our kinematic matching method ensures consistent walking patterns between real subjects and their virtual characters under diverse clothing. As detailed in Algorithm~\ref{alg:algorithm}, we first construct a local coordinate system for each bone from the original 3D pose and compute its rotation in quaternion form. These local-space rotations are then transferred to the target skeleton according to the predefined joint correspondence, enabling stable and gimbal-lock-free~\cite{alaimo2013comparison} reproduction of subject-specific motion. This mapping-driven alignment ensures faithful motion transfer under all clothing variations (Figure~\ref{fig2}c).

\vspace{-0.3cm}
\paragraph{Scene Construction.}
To ensure scene diversity, we collect 20 indoor and outdoor environments and position 8 cameras at a height of 2.5 m in each scene. The cameras are placed 45 degrees apart in a circle with a 4-meter radius, capturing humans from various angles, as shown in Figure~\ref{fig2}d. In addition to clothing variation, BarbieGait also introduces real-world factors such as \textbf{occlusion} and \textbf{lighting} changes by placing common obstacles (e.g., chairs, walls) and simulating day-to-night lighting conditions to enhance realism and robustness.

\begin{algorithm}[t]
\setstretch{0.8} 
\caption{Kinematic Motion Matching}
\label{alg:algorithm}
\textbf{Input}: Initial $A$ Pose, 3D Pose $P_{t}$ at frame $t$, $root$ is the root joint of the human \\
\textbf{Function} $CalQ$: Calculate the unit quaternion that parameterizes the rotational transformation of the local coordinate system with respect to the world coordinate system. Note that the local coordinate system is based on each bone and its parent bone.    \\
\textbf{Output}:\hbox{Animated Pose at frame $t$}
\begin{algorithmic}[1] 
\STATE $Q_{s} \gets CalQ(A)$ 
\STATE $Q_{t} \gets CalQ(P_{t})$
\FOR{$k$ in $bones$}
\IF{$bones[k]$ \textbf{is} $root$}  
\STATE $Q \gets Q_{s}(k) \times Q_{t}(k)$
\ELSE
\STATE $Q \gets Q_{s}(k) \times \Delta{Q_{t}(k_{p})} \times Q_{t}(k)$
\ENDIF 
\STATE Drive bone $bones[k]$ by rotating $Q$
\ENDFOR
\end{algorithmic}
\end{algorithm}
\vspace{-15pt}


\paragraph{Rendering.}
We use an NVIDIA GPU cluster for photorealistic rendering in Figure \ref{fig2}e. Unlike SynBody \cite{yang2023synbody}, which renders full 1920$\times$1080 images, we focus on human and shadow regions, rendering static areas only once. This approach enhances focus on the human region, simplifies post-processing, and boosts rendering speed by 5-6 times.

\vspace{-0.1cm}
\subsection{Data Processing}
To enable more comprehensive research and analysis, we process the rendered data to obtain multimodal data. In addition to the synthetic RGB data, we obtain segmented silhouettes using PaddleSeg \cite{liu2021paddleseg} and predict 2D human poses for each frame using HRNet \cite{wang2020deep}.

\subsection{Data Statistics and Evaluation Protocols}
\par BarbieGait comprises 521 subjects (174 male, 347 female) spanning a broad range of ages (5-80 years), heights (110-192 cm), and weights (15-115 kg), ensuring high diversity in body shape. Among them, 261 subjects are used for training (602,508 sequences) and 260 for testing (600,816 sequences). As shown in Table~\ref{tab1}, our dataset features one of the largest sequence counts and offers superior outfit diversity, with 100 clothing conditions per subject. 

\begin{figure}[t] 
    \centering
    \includegraphics[width=1.0\columnwidth]{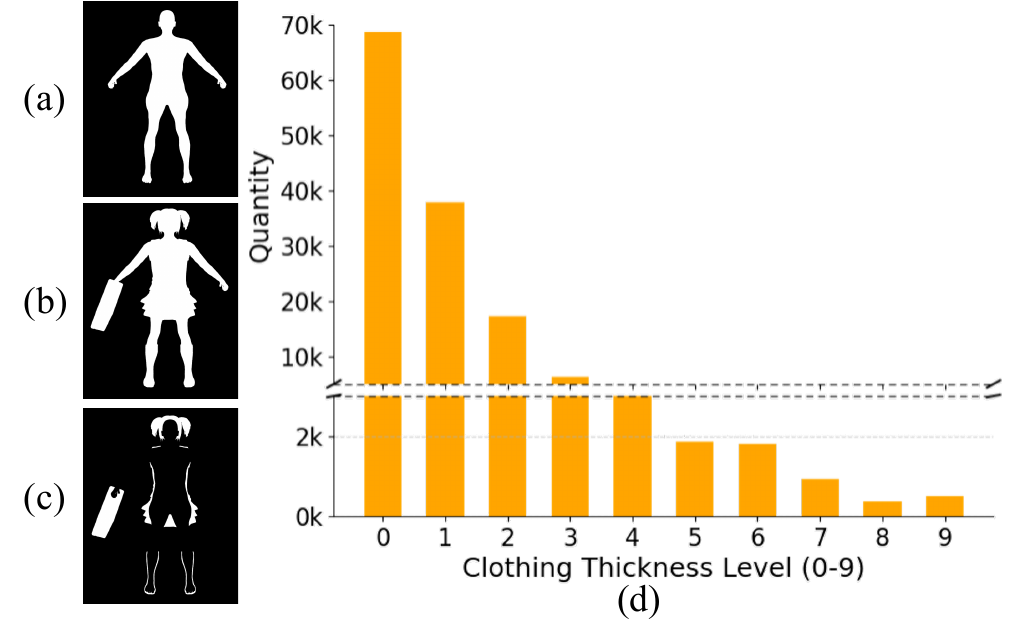}  
    \vspace{-0.9cm}
    \caption{Clothing Complexity and Thickness: (a) Silhouette without clothes. (b) Silhouette with clothes. (c) Non-overlapping area between (a) and (b) indicates garment complexity. (d) Distribution of subjects across thickness levels.}  
    \label{fig3}  
\end{figure}

\par We evaluate using Rank-1 accuracy (R1) and mean Average Precision (mAP). To analyze the effect of clothing thickness, we introduce a metric based on silhouette differences. Specifically, we render each subject’s binary silhouette without clothes (Figure~\ref{fig3}a) and with various outfits (Figure~\ref{fig3}b), then compute the non-overlapping region between them (Figure~\ref{fig3}c) as a measure of clothing complexity. This area is normalized by the unclothed silhouette to define relative clothing thickness. We then divide clothing into ten levels (THK0–THK9), each representing a 15\% increase in thickness. Figure~\ref{fig3}d shows the distribution across levels, highlighting the influence of clothing variation on gait recognition.

\subsection{Privacy Statement}
We obtain consent from all subjects for research use only. Raw data remains private, and only synthesized data will be released without any personal visual information such as faces, environments, or RGB images.

\section{GaitCLIF}
BarbieGait, as a versatile dataset, especially provides a solid foundation for addressing the challenge of gait recognition under cloth-changing conditions. To tackle this challenge and learn \emph{cloth-invariant} features under cloth-changing conditions, we explore solutions from two key perspectives: (1) removing cloth-specific statistics and (2) preserving fine-grained motion details. By combining these strategies, we introduce a straightforward yet powerful approach, \textbf{GaitCLIF} (Gait-oriented CLoth-Invariant Feature) for gait recognition in cloth-changing conditions.

\subsection{Removing Cloth-Specific Statistics}
Clothing-related statistics significantly contribute to large intra-class variance and form sub-domains associated with clothing-related features discrepancy, which hinders the model's ability to extract gait identity-specific information. Ideally, a gait recognition model for cloth-changing conditions should learn cloth-invariant features and minimize the impact of clothing diversity. 

 
\par Therefore, we view removal clothing-induced variations at each frame as a crucial step toward learning cloth-invariant features, preventing clothing-related variations from interfering with identity cues. To address this, we adopt normalization strategies designed to eliminate style-specific statistics from the feature channels. In particular, we find that the commonly used Instance Normalization in domain-invariant learning~\cite{pan2018two, jin2020style,choi2021meta,zhuang2020rethinking,chang2023learning,jiao2022dynamically} is not suitable for gait recognition due to noise in each channel of the silhouette features. Therefore, we propose \textit{Gait-Oriented Normalization} (GON), a method inspired by Layer Normalization (LN) but specifically designed to account for the characteristics of gait data. GON is effectively suited to remove cloth-specific statistics from features across channels for each frame, as illustrated in Figure~\ref{fig4}a. The details of GON will be introduced in the next subsection, combining our efforts to preserve fine-grained motion details.

\subsection{Preserving Fine-Grained Motion Details}
Fine-grained motion details are essential for recognizing individuals across various walking conditions~\cite{fan2020gaitpart, wang2023gaitparsing, huang2024occluded}. Thus, we identify them as another key factor in learning cloth-invariant features, especially under extensive clothing changes. To this end, we propose GON-P3D/GON-3D and GON-FC to preserve the motion details at both the frame-level and sequence-level in a simple yet effective manner.

\begin{figure}[t] 
    \centering
    \includegraphics[width=1.0\columnwidth]{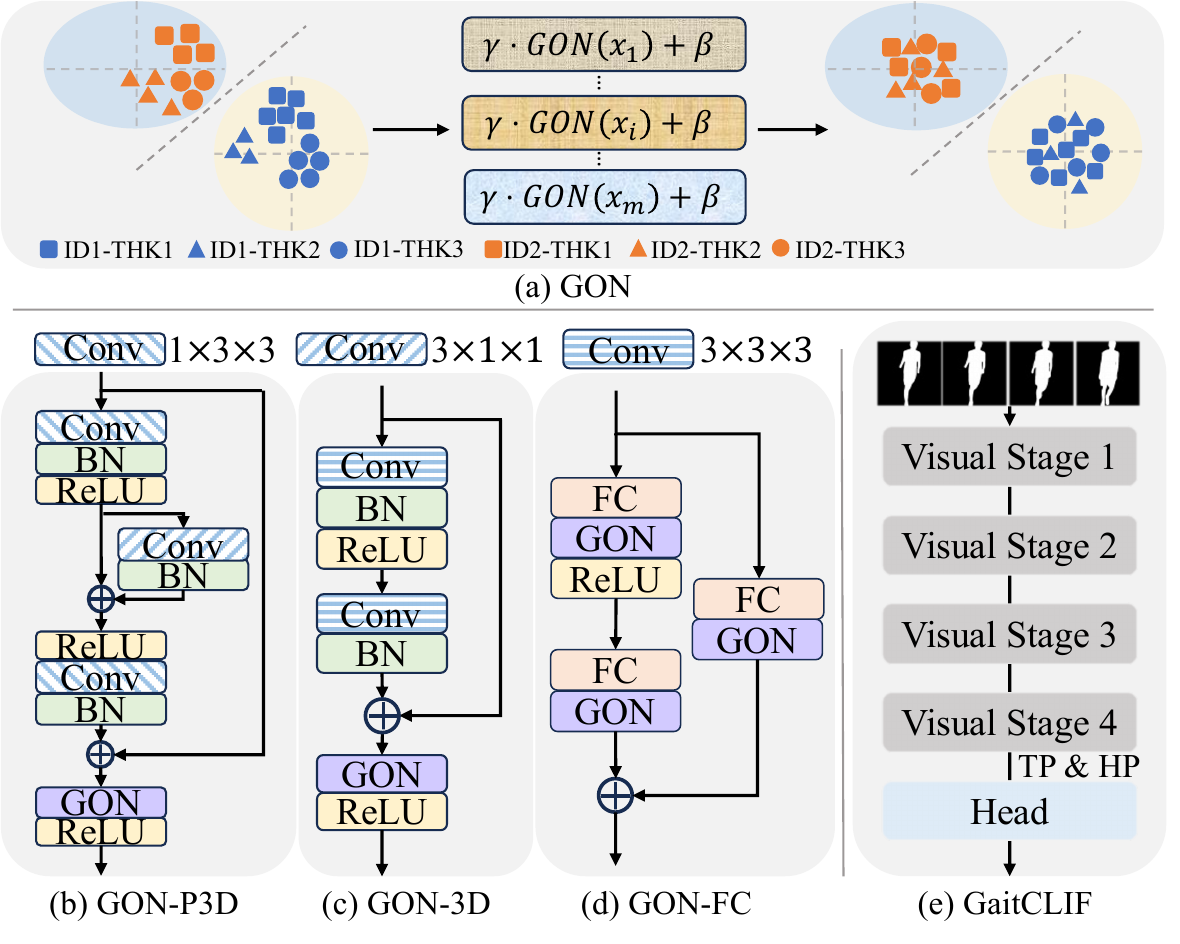}  
    \vspace{-0.6cm}
    \caption{Overview of GaitCLIF.
    (a) GON, the core normalization unit.
    (b) GON-P3D and (c) GON-3D, two GON-based visual blocks used in the visual stages of GaitCLIF.
    (d) GON-FC, a GON-enhanced FC block used in the Head of GaitCLIF.
    (e) The overall GaitCLIF framework for cross-clothing gait recognition.}  
    \label{fig4}  
\end{figure}

\vspace{-0.3cm}
\paragraph{Frame-Level Modelling}
\par Clothing variations are not uniform across different body regions of the whole feature $\text{X} \in \mathbb{R}^{N \times C \times H \times W}$. For example, the head has minimal changes, while the lower body is more variable due to different clothing styles like skinny pants, wide-leg pants, or skirts. Applying global normalization strategies fails to account for these fine-grained variations, making it harder for the model to learn consistent identity-related features across clothing changes.

\par Therefore, our GON module (Figure \ref{fig4}a) eliminates the impact of clothing variations based on the scale of changes present in the features of each horizontally segmented region ${x_0, \dots, x_i, \dots, x_m}$, thereby enhancing intra-class compactness for the same subject under different clothing conditions. The final clothing-invariant feature $\text{X}^{'}$ is produced by concatenating the individual segments features $\text{GON}(x_{i}) \in \mathbb{R}^{N \times C \times h_{i} \times W}$, which have been processed to alleviate fine-grained clothing variations and can be represented as:

\vspace{-0.5cm}
{
\begin{equation}
\text{X}^{'} = \text{GON}(\text{X})=\text{Cat}(\text{GON}(x_{0}), \dots, \text{GON}(x_{m}))
\end{equation}
}



\vspace{-0.5cm}
\begin{equation}
\text{GON}(x_{i}) = \gamma \left( \frac{x_{i} - \mu(x_{i})}{\sigma(x_{i})} \right) + \beta
\end{equation}
where $\mu(x_{i}), \sigma(x_{i})$ are the mean and standard deviation computed across all feature channels ($C$) and spatial dimensions ($h_{i}$, $W$) to minimize silhouette noise:

\vspace{-0.2cm}

\begin{equation}
\mu(x_{i}) = \frac{1}{Ch_{i}W} \sum_{c=1}^C \sum_{h=1}^{h_{i}} \sum_{w=1}^W x_{chw}
\end{equation}

\vspace{-0.4cm}

{\small
\begin{equation}
\sigma(x_{i}) = \sqrt{\frac{1}{Ch_{i}W} \sum_{c=1}^C \sum_{h=1}^{h_{i}} \sum_{w=1}^W \left( x_{chw} - \mu(x_{i}) \right)^2}
\end{equation}}
\vspace{-0.3cm}


\par To capture temporal dynamics in gait sequences, we extend GON with two temporal variants: GON-P3D and GON-3D (Figure~\ref{fig4}b, c). These modules apply temporal convolutions to enhance motion representation and improve frame-level clothing-invariant feature learning. 


\vspace{-0.4cm}
\paragraph{Sequence-Level Modelling}
\par We further enhance the model's capability to capture sequence-level cloth-invariant features after Temporal Pooling aggregation. The Separate Fully Connected (FC) layer used in the current mainstream gait recognition architecture \cite{fan2023opengait, fan2025opengait} is insufficient to deal with the extensive variations in clothing. Therefore, we focus on further enhancing the network's nonlinear expression capability for each fine-grained region and reduce clothing variance in the sequence-level. Specifically, GON-FC is a two-layer FC structure with GON applied after each FC layer, as illustrated in Figure \ref{fig4}d. This design further enhances the model's ability to express fine-grained motion patterns while introducing GON to improve the model's capacity to extract cloth-invariant information.




\begin{table}[t]
    \caption{Dataset-specific configurations including backbone and training settings.}
    \vspace{-0.2cm}
    \label{tab2}
    \centering
    \setlength{\tabcolsep}{0.5mm}
    \renewcommand\arraystretch{1.0}
    \fontsize{9}{9}\selectfont 
    \begin{tabular}{l|cccc}
    \toprule
    \textbf{Dataset}    & \textbf{Batch Size} & \textbf{Blocks}        & \textbf{Milestones}        & \textbf{Total} \\ \midrule
    BarbieGait & (8,8)      & {[}1,1,1,1{]} & (20k,40k,50k)   & 60k         \\
    CCPG \cite{li2023depth}       & (8,16)     & {[}1,1,1,1{]} & (20k,40k,50k)   & 60k         \\
    SUSTech1K \cite{shen2023lidargait}  & (8,8)      & {[}1,1,1,1{]} & (20k,40k,40k)   & 50k         \\
    Gait3D \cite{zheng2022gait}     & (32,4)     & {[}1,4,4,1{]} & (20k,40k,50k)   & 60k         \\
    GREW \cite{zhu2021gait}       & (32,4)     & {[}1,4,4,1{]} & (80k,120k,150k) & 180k        \\ \bottomrule 
    \end{tabular}
    \end{table}

\subsection{Overall Framework}


The overall GaitCLIF framework in Figure~\ref{fig4}e consists of four visual stages, temporal pooling (TP), horizontal pooling (HP), and a linear head for recognition. We further introduce two variants, \textit{GaitCLIF-P3D} and \textit{GaitCLIF-3D}, whose visual stages are built with \textit{GON-P3D} and \textit{GON-3D}, respectively, while both use \textit{GON-FC} as the head.

\section{Experiments}

Before conducting further experiments, we first assess the identity consistency between the real and synthetic data in BarbieGait after 3D pose matching and motion alignment. 
Since subject-specific gait identity is predominantly encoded in \emph{joint positions} and \emph{joint-angle} dynamics~\cite{winter1991biomechanics, whittle2014gait}, 
we quantify alignment quality along these two dimensions. 
\textbf{Our alignment is highly accurate:} the average joint position error is 12.2 mm (mainly due to hierarchical accumulated error) and the joint-angle error is only 0.02°, 
indicating precise spatial correspondence~\cite{zhang2021direct}. 

\subsection{Datasets}
\label{sec:datasets}
In our experiments, GaitCLIF not only serves as a robust baseline model for cross-clothing recognition in BarbieGait but also demonstrates consistent performance improvements on existing gait recognition benchmarks, such as two in-the-lab dataset CCPG \cite{li2023depth}, SUSTech1K \cite{shen2023lidargait} and two in-the-wild datasets Gait3D \cite{zheng2022gait} and GREW \cite{zhu2021gait}. 

 The detailed configurations, including the number of blocks in each visual stage, are summarized in Table~\ref{tab2}. All experiments follow the official evaluation protocols, with backbone settings aligned with prior works~\cite{li2023depth,shen2023lidargait,zheng2022gait}.

\begin{table*}[t]
    \caption{Performance comparison on BarbieGait when using predicted silhouette and 2D pose as input. Best result is in bold, and the second-best result is underlined. GaitCLIF-P3D$^{*}$ uses heatmaps as input, following the same data processing pipeline as SkeletonGait. In our experiments, THK0 serves as the gallery and THK1-THK9 as probes.}
    \setlength{\tabcolsep}{0.5mm}
    \fontsize{9}{9}\selectfont 
    \renewcommand{\arraystretch}{1.0} 

    \begin{tabular}{l|cccccccccccccccccc|cc}
        \toprule
        \multirow{2}{*}{\textbf{Methods}} & \multicolumn{2}{c}{\textbf{THK1}} & \multicolumn{2}{c}{\textbf{THK2}} & \multicolumn{2}{c}{\textbf{THK3}} & \multicolumn{2}{c}{\textbf{THK4}} & \multicolumn{2}{c}{\textbf{THK5}} & \multicolumn{2}{c}{\textbf{THK6}} & \multicolumn{2}{c}{\textbf{THK7}} & \multicolumn{2}{c}{\textbf{THK8}} & \multicolumn{2}{c|}{\textbf{THK9}} & \multicolumn{2}{c}{\textbf{AVG}} \\
                                          & \textbf{R1}     & \textbf{mAP}    & \textbf{R1}     & \textbf{mAP}    & \textbf{R1}     & \textbf{mAP}    & \textbf{R1}     & \textbf{mAP}    & \textbf{R1}     & \textbf{mAP}    & \textbf{R1}     & \textbf{mAP}    & \textbf{R1}     & \textbf{mAP}    & \textbf{R1}     & \textbf{mAP}    & \textbf{R1}      & \textbf{mAP}    & \textbf{R1}     & \textbf{mAP}   \\ \midrule
        GaitSet \cite{chao2019gaitset}                           & 20.6            & 23.3            & 15.8            & 18.9            & 12.7            & 15.9            & 12.3            & 15.5            & 7.4             & 10.9            & 6.0             & 8.9             & 5.3             & 7.8             & 4.1             & 7.5             & 3.2              & 6.2             & 9.7             & 12.8          \\
        GaitPart \cite{fan2020gaitpart}                          & 50.4            & 47.7            & 42.6            & 41.6            & 35.2            & 35.6            & 33.9            & 34.5            & 24.8            & 26.8            & 19.1            & 21.7            & 14.8            & 17.3            & 16.7            & 19.0            & 11.3             & 14.3            & 27.6            & 28.7          \\
        GaitGL \cite{lin2022gaitgl}                            & 44.7            & 34.0            & 37.5            & 29.4            & 31.1            & 25.2            & 31.2            & 25.7            & 22.3            & 19.5            & 18.3            & 16.6            & 13.0            & 13.2            & 14.8            & 14.6            & 13.8             & 13.8            & 25.2            & 21.3          \\
        GaitBase \cite{fan2023opengait}                          & 32.1            & 32.9            & 27.7            & 29.2            & 22.8            & 24.7            & 21.0            & 23.1            & 16.2            & 19.2            & 13.0            & 16.1            & 10.6            & 13.5            & 11.9            & 15.1            & 9.6              & 12.6            & 18.3            & 20.7          \\
        DeepGaitV2-2D \cite{fan2025opengait}                     & 59.6            & 57.9            & 51.6            & 51.3            & 46.1            & 45.5            & 42.6            & 43.7            & 36.4            & 38.2            & 29.3            & 31.9            & 22.5            & 25.5            & 25.5            & 28.7            & 18.0             & 21.9            & 36.8            & 38.3          \\
        DeepGaitV2-3D \cite{fan2025opengait}                     & 87.3            & 73.8            & 83.5            & 70.5            & 79.7            & 66.9            & 79.5            & 66.5            & 76.8            & 62.7            & 68.0            & 57.7            & 55.5            & 47.6            & 61.7            & 50.4            & 53.4            & 45.9            & 71.7            & 60.2          \\
        DeepGaitV2-P3D \cite{fan2025opengait}                    & 85.4            & 72.4            & 81.0            & 68.8            & 76.7            & 64.8            & 76.1            & 64.4            & 72.8            & 59.9            & 63.1            & 54.2            & 50.9            & 44.8            & 55.3            & 46.9            & 47.9             & 42.3            & 67.7            & 57.6          \\ \midrule
        \rowcolor{blue!10} GaitCLIF-P3D   (ours)             & \underline{88.1}     & \underline{74.2}     & \underline{84.8}     & \underline{71.5}     & \underline{82.0}     & \underline{68.3}     & \underline{82.3}     & \underline{68.5}     & \underline{80.1}     & \underline{65.1}     & \underline{71.6}     & \underline{60.5}     & \underline{62.8}     & \underline{53.5}     & \underline{67.5}     & \underline{55.1}     & \underline{61.1}      & \underline{51.9}     & \underline{75.6}     & \underline{63.2}    \\
        \rowcolor{blue!10} GaitCLIF-3D (ours)                & \textbf{90.7}   & \textbf{75.3}   & \textbf{88.1}   & \textbf{73.0}   & \textbf{85.8}   & \textbf{70.1}   & \textbf{85.6}   & \textbf{70.2}   & \textbf{84.5}   & \textbf{67.2}   & \textbf{77.8}   & \textbf{64.1}   & \textbf{69.2}   & \textbf{57.1}   & \textbf{73.8}   & \textbf{58.1}   & \textbf{68.5}    & \textbf{55.9}   & \textbf{80.4}   & \textbf{65.7} \\ \midrule\midrule
        GaitGraph \cite{teepe2021gaitgraph}                         & 14.0            & 18.0            & 12.3            & 16.6            & 11.8            & 16.0            & 11.0            & 15.5            & 9.8             & 13.6            & 9.3             & 13.3            & 6.6             & 10.6            & 8.2             & 11.8            & 6.5              & 10.3            & 10.0            & 14.0          \\
        GaitGraph2 \cite{teepe2022towards}                        & 36.5            & 37.2            & 33.5            & 34.9            & 32.6            & 34.4            & 30.8            & 33.2            & 28.9            & 31.4            & 24.1            & 26.5            & 20.2            & 23.6            & 22.0            & 25.5            & 20.8             & 23.8            & 27.7            & 30.1          \\
        GaitTR \cite{zhang2023spatial}                            & 63.3            & 59.0            & 59.2            & 55.9            & 58.7            & 55.9            & 58.1            & 55.5            & 54.3            & 52.1            & 47.9            & 46.4            & 41.8            & 42.5            & 41.6            & 42.1            & 36.1             & 37.1            & 51.2            & 49.6          \\
        GPGait \cite{fu2023gpgait}                            & 79.4            & 74.8            & 74.5            & 70.6            & 71.1            & 68.0            & 67.2            & 64.4            & 61.8            & 60.4            & 52.0            & 51.7            & 44.2            & 46.7            & 45.1            & 45.4            & 36.9             & 39.3            & 59.1            & 57.9          \\
        SkeletonGait \cite{fan2024skeletongait}                     & \underline{91.4}     & \underline{85.5}     & \underline{88.7}     & \underline{82.6}     & \underline{87.0}     & \underline{80.9}     & \underline{84.0}     & \underline{78.3}     & \underline{81.2}     & \underline{75.7}     & \underline{73.1}     & \underline{67.8}     & \underline{66.0}     & \underline{62.9}     & \underline{66.4}     & \underline{62.7}     & \underline{56.1}      & \underline{54.5}     & \underline{77.1}     & \underline{72.3}    \\ \midrule
        \rowcolor{blue!10} GaitCLIF-P3D$^{*}$ (ours)      & \textbf{92.1}   & \textbf{86.1}   & \textbf{89.1}   & \textbf{83.1}   & \textbf{87.5}   & \textbf{81.5}   & \textbf{85.1}   & \textbf{79.3}   & \textbf{82.3}   & \textbf{76.8}   & \textbf{74.3}   & \textbf{69.1}   & \textbf{68.6}   & \textbf{64.3}   & \textbf{67.5}   & \textbf{64.2}   & \textbf{56.7}    & \textbf{55.1}   & \textbf{78.1}   & \textbf{73.3} \\ \bottomrule
        \end{tabular}
    \vspace{-0.2cm}
    
    \label{tab3}
    \end{table*}
\begin{table}[t]
    \centering
    \caption{Ablation studies for GaitCLIF-P3D on BarbieGait.}
    \vspace{-0.2cm} 
    \setlength{\tabcolsep}{2.0mm}
    \renewcommand{\arraystretch}{1} 
    \fontsize{9}{8}\selectfont 
    \begin{tabular}{cc|cc}
        \toprule
        \textbf{GON-P3D} & \textbf{GON-FC} & \textbf{AVG-R1} (\%) & \textbf{AVG-mAP} (\%) \\
        \midrule
        $\times$ & $\times$ & 67.7 & 57.6 \\
        $\surd$ & $\times$ & 69.8 & 57.6 \\
        $\times$ & $\surd$ & 69.2 & 59.1 \\
        $\surd$ & $\surd$ & \textbf{75.6} & \textbf{63.2} \\
        \bottomrule
        \end{tabular}

    \label{tab4}
    \end{table}

\begin{table*}[ht]
\caption{Performance comparison on CCPG and SUSTech1K. For clarity, DeepGaitV2 and GaitCLIF refer to P3D-based models.}
\label{tab5}
\centering
\setlength{\tabcolsep}{1.1mm}
\renewcommand\arraystretch{1.0}
\fontsize{9}{9}\selectfont 
\begin{tabular}{l|cccccccc|cccccccccc}
\toprule
\multirow{3}{*}{\textbf{Methods}} & \multicolumn{8}{c|}{\textbf{CCPG}}                                                                                                       & \multicolumn{10}{c}{\textbf{SUSTech1K}}                                                                                                                                      \\
                       & \multicolumn{2}{c}{\textbf{CL}} & \multicolumn{2}{c}{\textbf{UP}} & \multicolumn{2}{c}{\textbf{DN}} & \multicolumn{2}{c|}{\textbf{MEAN}} & \textbf{NM}    & \textbf{BG}    & \textbf{CL}    & \textbf{CA}    & \textbf{UM}    & \textbf{UN}    & \textbf{OC}    & \textbf{NG}    & \multicolumn{2}{c}{\textbf{Overall}} \\ \cmidrule(lr){2-9} \cmidrule(lr){10-19}
                       & \textbf{R1}     & \textbf{mAP}    & \textbf{R1}     & \textbf{mAP}    & \textbf{R1}     & \textbf{mAP}    & \textbf{R1}     & \textbf{mAP}  & \multicolumn{8}{c}{\textbf{R1}}                                                                                                  & \textbf{R1}  & \textbf{R5} \\ \midrule
GaitSet\cite{chao2019gaitset}                & 77.7           & 46.4           & 83.5           & 59.6           & 83.2           & 61.4           & 81.5             & 55.8           & 69.1          & 68.2          & 37.4          & 65.0          & 63.1          & 61.0          & 67.2          & 23.0          & 65.0             & 84.8            \\
GaitPart\cite{fan2020gaitpart}               & 77.8           & 45.5           & 84.5           & 63.1           & 83.3           & 60.1           & 81.9             & 56.2           & 62.2          & 62.8          & 33.1          & 59.5          & 57.2          & 54.8          & 57.2          & 21.7          & 59.2             & 80.8            \\
GaitBase\cite{fan2023opengait}               & 88.5           & \textemdash               & 92.7       & \textemdash    & 93.4           & \textemdash    & 91.5             & \textemdash                & 81.5          & 77.5          & 49.6          & 75.8          & 75.5          & 76.7          & 81.4          & 25.9          & 76.1             & 89.4            \\
SkeletonGait~\cite{fan2024skeletongait}             & 52.4           & 20.8           & 65.4           & 35.8           & 72.8           & 40.3           & 63.5            & 32.3            & 67.9          & 63.5          & 36.5          & 61.6          & 58.1          & 67.2          & 79.1          & 50.1          & 63.0              & 83.5             \\
DeepGaitV2\cite{fan2025opengait}             & 90.5           & 63.6           & 95.3           & 81.1           & 92.9           & 79.4           & 92.9             & 74.7           & 87.4          & 84.1          & 53.4          & 81.3          & 86.1          & 84.8          & 88.5          & 28.8          & 82.3             & 92.5            \\ \midrule
\rowcolor{blue!10} GaitCLIF (ours)         & \textbf{93.4}  & \textbf{66.4}  & \textbf{97.4}  & \textbf{84.3}  & \textbf{93.5}  & \textbf{80.3}  & \textbf{94.8}    & \textbf{77.0}  & \textbf{89.2} & \textbf{86.6} & \textbf{58.0} & \textbf{84.0} & \textbf{87.7} & \textbf{88.0} & \textbf{90.7} & \textbf{29.9} & \textbf{84.7}    & \textbf{93.6}   \\ \bottomrule 
\end{tabular}
\vspace{-0.4cm}
\end{table*}

\begin{table}[t]
\caption{Performance comparison on Gait3D and GREW. For clarity, DeepGaitV2 and GaitCLIF refer to P3D-based models.}
\label{tab6}
\centering
\setlength{\tabcolsep}{2.5mm}
\renewcommand\arraystretch{1.0}
\fontsize{9}{9}\selectfont 
\begin{tabular}{l|ccc|cc}
\toprule
\multirow{2}{*}{\textbf{Methods}}  & \multicolumn{3}{c|}{\textbf{Gait3D}}             & \multicolumn{2}{c}{\textbf{GREW}} \\ \cmidrule(lr){2-4} \cmidrule(lr){5-6}
                        & \textbf{R1}            & \textbf{R5}            & \textbf{mAP}            & \textbf{R1}             & \textbf{R5}             \\ \midrule
GaitSet\cite{chao2019gaitset}                 & 36.7          & 58.3          & 30.0          & 46.3           & 63.6           \\
GaitPart\cite{fan2020gaitpart}                & 28.2          & 47.6          & 21.6          & 44.0           & 60.7           \\
GaitBase\cite{fan2023opengait}                & 60.1          & \textemdash               & \textemdash               & 64.6           & \textemdash                \\
GaitMoE~\cite{huang2024occluded}                  & 73.7          & \textemdash            & 66.2          & 79.6            & 89.1            \\
DeepGaitV2\cite{fan2025opengait}              & 74.4          & 88.0          & 65.8          & 77.7           & 88.9           \\ \midrule
\rowcolor{blue!10} GaitCLIF (ours) & \textbf{76.5} & \textbf{88.5} & \textbf{67.9} & \textbf{80.2}  & \textbf{89.2}  \\ \bottomrule 
\end{tabular}
\end{table}

\subsection{Performance on BarbieGait}
\paragraph{Baseline Performance.}
BarbieGait opens new directions for gait recognition. As a first step, we input Blender-rendered ideal silhouettes into DeepGaitV2-P3D \cite{fan2025opengait} to fully exploit this modality. The results show that the model achieves high accuracy under ideal conditions (R1: 91.2\%, mAP: 83.4\%) despite 100 outfit variations per subject. In contrast, real segmented silhouettes with noise cause a notable drop (R1: 67.7\%, mAP: 57.6\%), highlighting both the challenges and potential of cloth-changing scenarios.
\par We compare mainstream appearance-based and pose-based methods on BarbieGait in Table~\ref{tab3}, leading to the following observations:
(1) Appearance-based methods have advanced significantly—from GaitSet to the more robust DeepGaitV2 series. The 2D, 3D, and P3D variants of DeepGaitV2 demonstrate the importance of temporal modeling, especially under clothing variation, as joint dynamics and clothing motion over time capture key gait cues. Our GaitCLIF further boosts the overall performance, with GaitCLIF-P3D and GaitCLIF-3D reaching 63.2\% and 65.7\% mAP, respectively.
(2) Pose-based methods are naturally robust to appearance changes, but their advantages under clothing variation remain underexplored due to limited data diversity. BarbieGait provides abundant cross-clothing pose data, enabling a clearer evaluation of their effectiveness. With heatmaps as input, SkeletonGait achieves an mAP to 72.3\%, surpassing appearance-based GaitCLIF-3D (mAP: 65.7\%). 
Further using GaitCLIF-P3D as the backbone gives the best heatmap-based result (mAP: 73.3\%).

\vspace{-0.5cm}
\paragraph{Ablation Studies.}
We conduct ablation studies to analyze the roles of GON-P3D and GON-FC, with results shown in the first part of Table~\ref{tab4}. GON-P3D improves frame-level modeling by suppressing cloth-induced fluctuations, while GON-FC enhances sequence-level aggregation and stabilizes identity cues. Their combination yields the best performance, confirming their complementary contributions to learning cloth-invariant gait features.

\subsection{Performance on Real Datasets}

To further validate GaitCLIF's robustness, we evaluate it on four diverse real-world datasets: CCPG, SUSTech1K, Gait3D, and GREW. Though originally designed for clothing variation, GaitCLIF also mitigates part-level appearance changes from view shifts or carrying (e.g., bags), supporting its generalization. As shown in Table~\ref{tab5}, GaitCLIF improves accuracy across all clothing conditions on CCPG, with R1 and mAP gains of 1.9\% and 2.3\% under the ReID protocol~\cite{li2023depth}. On SUSTech1K, it boosts R1 and R5 by 2.4\% and 1.1\%. For in-the-wild datasets such as Gait3D and GREW, where clothing variation is limited, direct application may cause excessive intra-class divergence. To address this, we use only GON-FC to enhance nonlinear mapping and cloth-invariant feature learning. Table~\ref{tab6} shows that GaitCLIF achieves 76.5\% R1 / 67.9\% mAP on Gait3D and 80.2\% R1 / 89.2\% R5 on GREW.

\begin{table}[t]
\caption{Cross-domain performance on CCPG using different training strategies. ``Scratch'' indicates training only on CCPG, while ``Pretrain'' denotes BarbieGait pretraining followed by CCPG fine-tuning. GaitCLIF refer to P3D-based models.}
\label{tab7}
\centering
\setlength{\tabcolsep}{0.7mm}
\fontsize{9}{9}\selectfont 

\begin{tabular}{l|cccccccc}
    \toprule
    \multirow{3}{*}{\textbf{Model}}   & \multicolumn{8}{c}{\textbf{CCPG}}                                                                                                       \\
                             & \multicolumn{2}{c}{\textbf{CL}} & \multicolumn{2}{c}{\textbf{UP}} & \multicolumn{2}{c}{\textbf{DN}} & \multicolumn{2}{c}{\textbf{MEAN}} \\ \cmidrule{2-9} 
                             & \multicolumn{8}{c}{\textbf{R1 \& mAP }}                                                                                        \\ \midrule
    DeepGaitV2               & 90.5           & 63.6           & 95.3           & 81.1           & 92.9           & 79.4           & 92.9            & 74.7           \\
    GaitCLIF (Scratch)
    & 93.4           & 66.4           & 97.4           & 84.3           & 93.5           & 80.3           & 94.8            & 77.0           \\
    \rowcolor{blue!10} GaitCLIF (Pretrain)
    & \textbf{93.9} & \textbf{69.8} & \textbf{97.9} & \textbf{85.6} & \textbf{95.7} & \textbf{83.0} & \textbf{95.8}  & \textbf{79.5}  \\ \bottomrule 
    \end{tabular}

\end{table}

\subsection{BarbieGait-to-Real Evaluation}
\paragraph{Downstream to Real.}
BarbieGait, with its rich clothing variations, also demonstrates promising cross-domain generalization to real-world datasets. When pretrained on BarbieGait and fine-tuned on real-world target domains with a small learning rate, the model outperforms training from scratch. For instance, on the cloth-changing dataset CCPG, our model achieves R1 / mAP of 95.8\% / 79.5\% with BarbieGait pretraining, compared to 94.8\% / 77.0\% when trained solely on the target data. These results in Table~\ref{tab7} validate the effectiveness of BarbieGait as a universal pretraining source for real-world gait recognition.

\vspace{-0.3cm}
\paragraph{Upstream to Real.}
We further extend our exploration of BarbieGait by investigating its potential impact on upstream human pose estimation tasks~\cite{wang2020deep,xu2022vitpose}. A unique property of BarbieGait is that the same subject is rendered in diverse clothing conditions while preserving consistent identity information, with each image accurately paired with its corresponding 2D pose. This identity-consistent yet clothing-diverse setting provides an ideal supervision source for learning pose representations that are less affected by appearance changes. We select keypoints consistent with COCO~\cite{lin2014microsoft} in both semantics and quantity (covering head, trunk, and lower limbs), and retrain ViTPose~\cite{xu2022vitpose} on BarbieGait to adapt the model for gait-specific scenarios. The  implementation details are provided in the supplementary material.
As shown in Table~\ref{tab8}, our method achieves the best performance on the CCPG dataset, with an average R1 accuracy of 83.0\% and mAP of 51.5\%, surpassing SkeletonGait by +19.5\% and +19.2\%, and DPGait by +3.9\% in R1. Notably, while DPGait~\cite{lang2025beyond} also improves upstream pose estimation, it relies on large-scale motion data for training without enforcing identity-consistent constraints. In contrast, our BarbieGait-based approach leverages identity-preserving supervision across clothing variations, yielding more discriminative and robust gait representations.


\begin{table}[t]
\caption{BarbieGait improves upstream pose estimation, enabling better generalization to real-world dataset CCPG.}
\label{tab8}
\centering
\setlength{\tabcolsep}{0.9mm}
\renewcommand\arraystretch{1.0}
\fontsize{9}{9}\selectfont 
\begin{tabular}{l|cccccccc}
    \toprule
    \multirow{3}{*}{\textbf{Methods}} & \multicolumn{8}{c}{\textbf{CCPG}}                                                                                                       \\
                                      & \multicolumn{2}{c}{\textbf{CL}} & \multicolumn{2}{c}{\textbf{UP}} & \multicolumn{2}{c}{\textbf{DN}} & \multicolumn{2}{c}{\textbf{MEAN}} \\ \cmidrule{2-9}  
                                      & \multicolumn{8}{c}{\textbf{R1 \& mAP}}                                                                                                   \\ \midrule
    SkeletonGait~\cite{fan2024skeletongait}                      & 52.4           & 20.8           & 65.4           & 35.8           & 72.8           & 40.3           & 63.5            & 32.3            \\
    DPGait~\cite{lang2025beyond}                            & 70.7           &  \textemdash              & 82.4           & \textemdash               & 84.2           & \textemdash               & 79.1            & \textemdash                \\ \midrule
    \rowcolor{blue!10} Ours                              & \textbf{76.9}  & \textbf{40.0}  & \textbf{84.9}  & \textbf{56.7}  & \textbf{87.2}  & \textbf{57.7}  & \textbf{83.0}   & \textbf{51.5}   \\ \bottomrule
    \end{tabular}
\end{table}

\section{Conclusion}
In conclusion, this paper introduces BarbieGait, an identity-consistent synthetic dataset designed to address two long-standing limitations in cloth-changing gait recognition: the scarcity of extensive clothing variation and the difficulty of preserving subject-specific identity cues in synthetic data. Building upon this dataset, we further demonstrate the potential of cloth-changing gait recognition by developing GaitCLIF, a strong baseline trained on BarbieGait that achieves consistent improvements on both BarbieGait and existing gait benchmarks. In addition, our exploratory studies show that BarbieGait also serves as an effective pretraining source for both downstream gait recognition and upstream human pose estimation.

\section{Acknowledgement}
This work is jointly supported by Joint Fund for the Provincial Science and Technology R\&D Program of Henan Province (245200810009), National Natural Science Foundation of China (62476027, 62276025) and the Fundamental Research Funds for the Central Universities (2253200026).

{
    \small
    \bibliographystyle{ieeenat_fullname}
    \bibliography{main}
}

\clearpage
\setcounter{page}{1}
\maketitlesupplementary
\setcounter{page}{1}
\setcounter{section}{6}
\setcounter{table}{8}
\setcounter{figure}{4}
\setcounter{equation}{4}

\section{Details of Kinematic Motion Matching}
Our kinematic matching process begins by extracting bone rotations from the input 3D keypoint sequence. Since the source 3D pose data only provides independent joint positions without a pre-defined articulated structure (i.e., no rigged skeleton or skinning rig), we construct a local coordinate system for each ``virtual bone segment’’ (i.e., a conceptual bone defined by two joints) based on joint-to-joint geometric relationships, enabling the extraction of pose parameters required for motion retargeting.

Specifically, for a bone defined by a parent joint $J_p$ and a child joint $J_c$, its primary axis is given by the vector $v = J_c - J_p$. To resolve the inherent twist ambiguity around this axis, we form a reference plane using adjacent joints to obtain a stable and reproducible secondary axis direction. This procedure uniquely determines the local coordinate system of each bone.

Based on these dynamically constructed local coordinate systems, we compute the world-space rotation quaternion of each bone for both the source and target skeletons, denoted as $Q_s$ and $Q_t$ in Algorithm 1. By further composing the child bone's world rotation with the inverse world rotation of its parent, we obtain the local rotation $\Delta Q_t(k_p)$, which characterizes the hierarchical relative motion of the source pose. All local rotations are then assembled according to the skeletal topology to form a complete hierarchical pose representation. Finally, these local rotations are applied to the target skeleton to accomplish the retargeting process via hierarchical bone rotation.

A more complete treatment of the analytical quaternion computation and the Blender implementation details underlying this kinematic matching pipeline can be found in~\cite{blender_animation_retargeting,Rokoko}.

\section{More Cloth-Changing Experiments}
We additionally evaluate GaitCLIF on established cloth-changing benchmarks, including HybridGait~\cite{dong2024hybridgait}, OU-ISIR~\cite{Makihara_CVATN2012}, CCVID~\cite{gu2022clothes}, and MEVID~\cite{davila2023mevid}. As shown in Table~\ref{tab_more_cloth}, GaitCLIF consistently improves performance under cloth-changing settings over DeepGaitV2 across both gait recognition and video Re-identification benchmarks. Specifically, it gains of 4.77 and 3.50 points on HybridGait and OU-ISIR, respectively, and further improves performance by 3.90 and 4.11 points on CCVID and MEVID, respectively. For gait recognition on HybridGait and OU-ISIR, we follow OpenGait~\cite{fan2023opengait}, using $64\times44$ silhouettes and sampling 30 frames from each sequence during training. For video Re-identification on CCVID and MEVID, we follow CCVID~\cite{gu2022clothes}, using $128\times88$ RGB images and sampling 8 frames from each sequence during training.

\begin{figure}[t] 
    \centering
    \includegraphics[width=0.95\columnwidth]{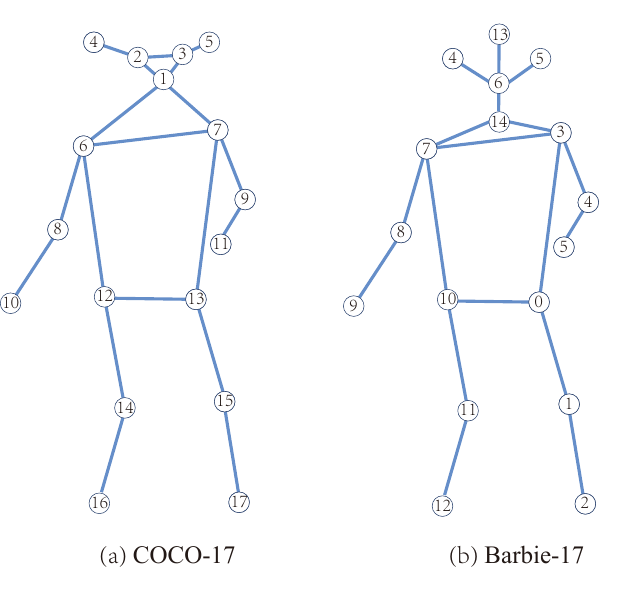}
    \caption{The pose format we used in our experiments. (a) COCO-17 format (b)Barbie-17 format. }  
    \label{fig5}  
\end{figure}

\begin{table}[t]
    \centering
    \caption{Performance comparison of DeepGaitV2 and GaitCLIF on additional cloth-changing benchmarks.}
    \label{tab_more_cloth}
    \setlength{\tabcolsep}{3.5pt}
    \resizebox{\columnwidth}{!}{%
    \begin{tabular}{c|cc|cc}
        \toprule
        \multicolumn{1}{c|}{\multirow{2}{*}{\textbf{Methods}}} & \multicolumn{2}{c|}{\textbf{Gait Recognition}} & \multicolumn{2}{c}{\textbf{Video Re-ID}} \\
        \cmidrule{2-5}
                   & \textbf{HybridGait}                           & \textbf{OU-ISIR} & \textbf{CCVID} & \textbf{MEVID} \\ \midrule
        DeepGaitV2~\cite{fan2025opengait} & 55.34                                          & 90.54            & 93.18          & 69.94          \\
        GaitCLIF (ours)   & \textbf{60.11}                                 & \textbf{94.04}   & \textbf{97.08} & \textbf{74.05} \\ \bottomrule
    \end{tabular}%
    }
\end{table}

\begin{table*}[t]
\setlength{\tabcolsep}{0.6mm}
\fontsize{9}{9}\selectfont 
\renewcommand{\arraystretch}{1.1} 

\begin{tabular}{cc|cccccccccccccccccc|cc}
    \toprule
    \multicolumn{1}{c}{\multirow{2}{*}{\textbf{GON-P3D}}} & \multicolumn{1}{c}{\multirow{2}{*}{\textbf{GON-FC}}} & \multicolumn{2}{c}{\textbf{THK1}}                                  & \multicolumn{2}{c}{\textbf{THK2}}                                  & \multicolumn{2}{c}{\textbf{THK3}}                                  & \multicolumn{2}{c}{\textbf{THK4}}                                  & \multicolumn{2}{c}{\textbf{THK5}}                                  & \multicolumn{2}{c}{\textbf{THK6}}                                  & \multicolumn{2}{c}{\textbf{THK7}}                                  & \multicolumn{2}{c}{\textbf{THK8}}                                  & \multicolumn{2}{c}{\textbf{THK9}}                                  & \multicolumn{2}{c}{\textbf{AVG}}                          \\
    \multicolumn{1}{c}{}                                  & \multicolumn{1}{c}{}                                 & \multicolumn{1}{c}{\textbf{R1}} & \multicolumn{1}{c}{\textbf{mAP}} & \multicolumn{1}{c}{\textbf{R1}} & \multicolumn{1}{c}{\textbf{mAP}} & \multicolumn{1}{c}{\textbf{R1}} & \multicolumn{1}{c}{\textbf{mAP}} & \multicolumn{1}{c}{\textbf{R1}} & \multicolumn{1}{c}{\textbf{mAP}} & \multicolumn{1}{c}{\textbf{R1}} & \multicolumn{1}{c}{\textbf{mAP}} & \multicolumn{1}{c}{\textbf{R1}} & \multicolumn{1}{c}{\textbf{mAP}} & \multicolumn{1}{c}{\textbf{R1}} & \multicolumn{1}{c}{\textbf{mAP}} & \multicolumn{1}{c}{\textbf{R1}} & \multicolumn{1}{c}{\textbf{mAP}} & \multicolumn{1}{c}{\textbf{R1}} & \multicolumn{1}{c}{\textbf{mAP}} & \multicolumn{1}{c}{\textbf{R1}} & \multicolumn{1}{c}{\textbf{mAP}} \\ \midrule
    $\times$                                  & $\times$                                            & 85.4                   & 72.4                    & 81.0                   & 68.8                    & 76.7                   & 64.8                    & 76.1                   & 64.4                    & 72.8                   & 59.9                    & 63.1                   & 54.2                    & 50.9                   & 44.8                    & 55.3                   & 46.9                    & 47.9                   & 42.3                     & 67.7                   & 57.6                    \\
    $\surd$                                      & $\times$                                            & 86.1                   & 71.3                    & 81.8                   & 68.0                    & 78.3                   & 64.3                    & 78.2                   & 64.2                    & 75.3                   & 59.8                    & 64.7                   & 54.2                    & 53.6                   & 45.8                    & 58.9                   & 47.2                    & 51.2                   & 43.6                     & 69.8                   & 57.6                    \\
    $\times$                                     & $\surd$                                            & 83.8                   & 71.7                    & 80.1                   & 68.6                    & 76.8                   & 65.2                    & 76.9                   & 65.2                    & 73.1                   & 60.7                    & 65.8                   & 56.6                    & 53.4                   & 46.8                    & 59.7                   & 50.3                    & 53.3                   & 46.4                     & 69.2                   & 59.1                    \\
    $\surd$                                      & $\surd$                                            & \textbf{88.1}                   & \textbf{74.2}                    & \textbf{84.8}                   & \textbf{71.5}                    & \textbf{82.0}                   & \textbf{68.3}                    & \textbf{82.3}                   & \textbf{68.5}                    & \textbf{80.1}                   & \textbf{65.1}                    & \textbf{71.6}                   & \textbf{60.5}                    & \textbf{62.8}                   & \textbf{53.5}                    & \textbf{67.5}                   & \textbf{55.1}                    & \textbf{61.1}                   & \textbf{51.9}                     & \textbf{75.6}                   & \textbf{63.2}                    \\ \bottomrule
    \end{tabular}

\caption{Ablation study of each module under different clothing conditions (THK1-THK9). We report Rank-1 (R1) accuracy and mean Average Precision (mAP) for each variant.}

\label{tab10}
\end{table*}

\begin{table*}[t]
\setlength{\tabcolsep}{0.95mm}
\fontsize{9}{9}\selectfont 
\renewcommand{\arraystretch}{1.1} 

\begin{tabular}{c|cccccccccccccccccc|cc}
    \toprule
    \multirow{2}{*}{\begin{tabular}[c]{@{}c@{}}\textbf{Norm} \\ \textbf{Type}\end{tabular}} & \multicolumn{2}{c}{\textbf{THK1}} & \multicolumn{2}{c}{\textbf{THK2}} & \multicolumn{2}{c}{\textbf{THK3}} & \multicolumn{2}{c}{\textbf{THK4}} & \multicolumn{2}{c}{\textbf{THK5}} & \multicolumn{2}{c}{\textbf{THK6}} & \multicolumn{2}{c}{\textbf{THK7}} & \multicolumn{2}{c}{\textbf{THK8}} & \multicolumn{2}{c}{\textbf{THK9}} & \multicolumn{2}{c}{\textbf{AVG}} \\
    & \textbf{R1}     & \textbf{mAP}    & \textbf{R1}     & \textbf{mAP}    & \textbf{R1}     & \textbf{mAP}    & \textbf{R1}     & \textbf{mAP}    & \textbf{R1}     & \textbf{mAP}    & \textbf{R1}     & \textbf{mAP}    & \textbf{R1}     & \textbf{mAP}    & \textbf{R1}     & \textbf{mAP}    & \textbf{R1}     & \textbf{mAP}    & \textbf{R1}    & \textbf{mAP}        \\ \midrule
    BN                           & 83.2        & 71.8       & 78.6        & 68.0       & 74.2        & 63.8       & 73.6        & 63.2       & 69.1        & 58.3       & 60.9        & 53.6       & 47.8        & 43.0       & 54.7        & 46.9       & 45.9        & 41.6        & 65.3       & 56.7       \\
    IN                           & 75.4        & 67.2       & 71.0        & 63.7       & 66.2        & 59.9       & 65.1        & 58.9       & 61.0        & 55.5       & 52.3        & 49.1       & 43.0        & 41.6       & 46.6        & 44.3       & 38.0        & 37.5       & 57.6       & 53.1       \\
    LN                           & 76.9        & 61.7       & 71.9        & 58.4       & 66.7        & 54.6       & 66.3        & 54.3       & 61.9        & 50.4       & 53.5        & 45.8       & 43.3        & 38.4       & 50.1        & 42.0       & 41.1        & 35.8       & 59.1       & 49.0       \\
    GON                          & \textbf{88.1}                   & \textbf{74.2}                    & \textbf{84.8}                   & \textbf{71.5}                    & \textbf{82.0}                   & \textbf{68.3}                    & \textbf{82.3}                   & \textbf{68.5}                    & \textbf{80.1}                   & \textbf{65.1}                    & \textbf{71.6}                   & \textbf{60.5}                    & \textbf{62.8}                   & \textbf{53.5}                    & \textbf{67.5}                   & \textbf{55.1}                    & \textbf{61.1}                   & \textbf{51.9}                     & \textbf{75.6}                   & \textbf{63.2}       \\ \bottomrule
    \end{tabular}

\caption{ Comparison of common normalization methods (BN, IN, LN) and our proposed GON across different clothing thickness levels.}

\label{tab11}
\end{table*}
\begin{figure*}[h] 
    \centering
    \includegraphics[width=1.0\textwidth]{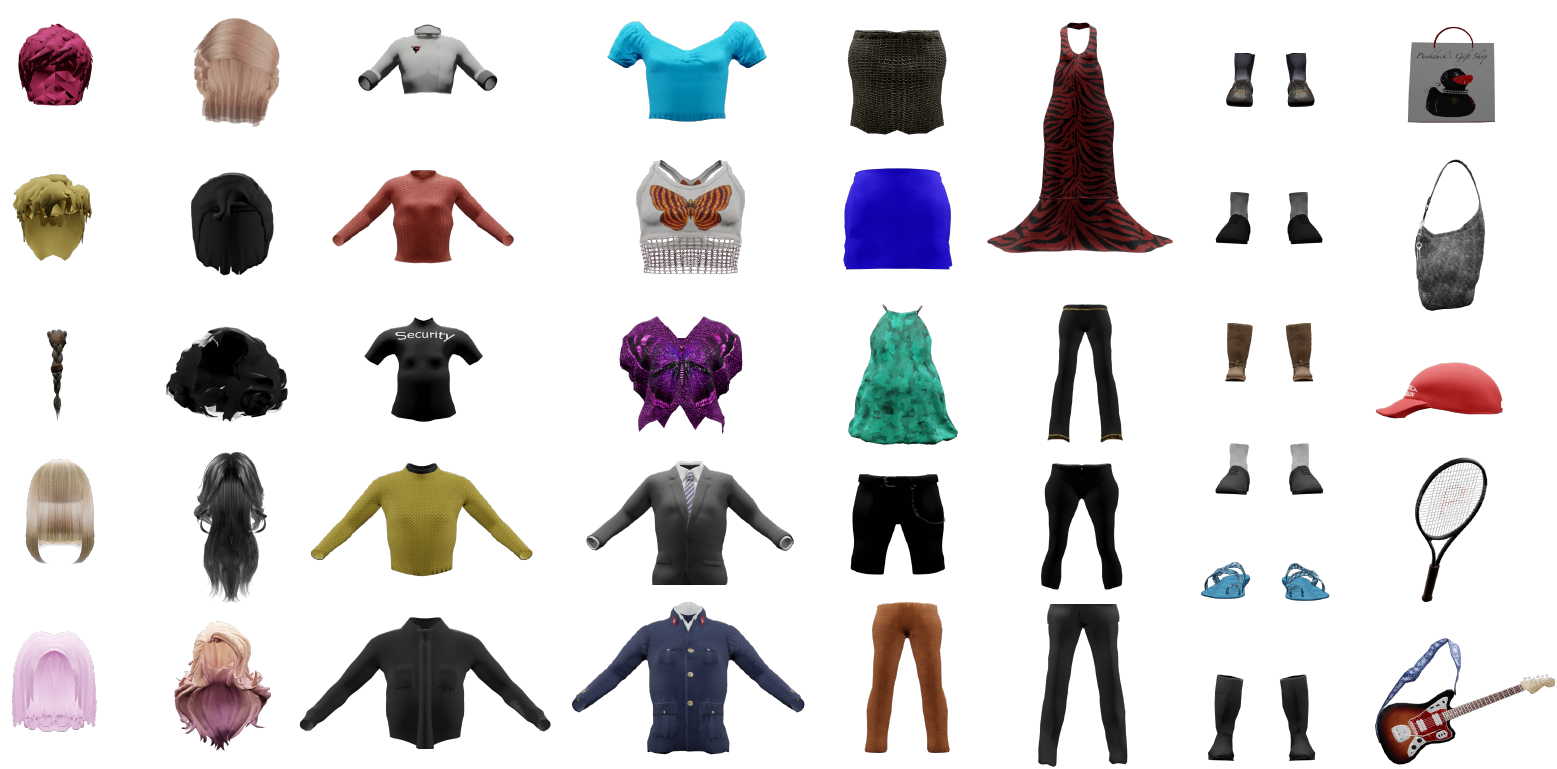}
    \caption{The Illustration of our diverse clothing. BarbieGait includes a variety of hairstyles, clothing, shoes, and carried objects, introducing significant clothing variations for gait recognition under cloth-changing conditions. }  
    \label{fig6}  
\end{figure*}
\begin{table*}[t]
\centering
\setlength{\tabcolsep}{0.45mm}
\renewcommand\arraystretch{1.1}
\fontsize{9}{9}\selectfont 
\begin{tabular}{ccc|cccccccc|cccccccccc}
    \toprule
    \multirow{3}{*}{\textbf{Methods}} & \multirow{3}{*}{\begin{tabular}[c]{@{}c@{}}\textbf{Pose} \\ \textbf{Type}\end{tabular}} & \multirow{3}{*}{\begin{tabular}[c]{@{}c@{}}\textbf{Upstream} \\ \textbf{Model}\end{tabular}} & \multicolumn{8}{c}{\textbf{CCPG}}                                                                                                       & \multicolumn{10}{c}{\textbf{SUSTech1K}}                                                                                                              \\
    &                            &                                 & \multicolumn{2}{c}{\textbf{CL}} & \multicolumn{2}{c}{\textbf{UP}} & \multicolumn{2}{c}{\textbf{DN}} & \multicolumn{2}{c|}{\textbf{Mean}} & \textbf{NM} & \textbf{BG} & \textbf{CL} & \textbf{CA} & \textbf{UM} & \textbf{UN} & \textbf{OC} & \textbf{NG} & \multicolumn{2}{c}{\textbf{Overall}} \\ \cmidrule{4-21} 
    &                            &                                 & \multicolumn{8}{c|}{\textbf{R1 \& mAP}}                                                                                                   & \multicolumn{8}{c}{\textbf{R1}}                                                                               & \textbf{R1}       & \textbf{R5}      \\ \midrule
    GaitTR~\cite{zhang2023spatial}                   & COCO-17                    & HRNet~\cite{wang2020deep}                           & 24.3           & 9.7            & 28.7           & 16.1           & 31.1           & 16.4           & 28.0             & 14.1            & 33.3        & 31.5        & 21.0        & 30.4        & 22.7        & 34.6        & 44.9        & 23.5        & 72.6              & 56.0             \\
    GaitGraph~\cite{teepe2021gaitgraph}                & COCO-17                    & HRNet~\cite{wang2020deep}                           & 5.0            & 2.4            & 5.7            & 4.0            & 7.3            & 4.2            & 6.0              & 3.5             & 22.2        & 18.2        & 6.8         & 18.6        & 13.4        & 19.2        & 27.3        & 16.4        & 18.6              & 40.2             \\
    SkeletonGait~\cite{fan2024skeletongait}             & COCO-17                    & HRNet~\cite{wang2020deep}                           & 52.4           & 20.8           & 65.4           & 35.8           & 72.8           & 40.3           & 63.5             & 32.3            & 55.0        & 51.0        & 24.7        & 49.9        & 42.3        & 52.0        & 62.8        & 43.9        & 50.1              & 72.6             \\
    DPGait~\cite{lang2025beyond}                   & COCO-17                    & ViTPose~\cite{xu2022vitpose}                         & 70.7           &  \textemdash              & 82.4           &  \textemdash              & 84.2           &  \textemdash              & 79.1             & \textemdash                & \textemdash            & \textemdash            & \textemdash            & \textemdash            & \textemdash            & \textemdash            &  \textemdash           & \textemdash            &  \textemdash                 & \textemdash                 \\ \midrule
    Ours                     & Barbie-17                  & ViTPose~\cite{xu2022vitpose}                         & \textbf{76.9}           & \textbf{40.0}           & \textbf{84.9}           & \textbf{56.7}           & \textbf{87.2}           & \textbf{57.7}           & \textbf{83.0}             & \textbf{51.5}            & \textbf{64.3}        & \textbf{50.6}        & \textbf{39.9}        & \textbf{59.0}        & \textbf{52.6}        & \textbf{61.9}        & \textbf{69.8}        & \textbf{40.1}        & \textbf{59.7}              & \textbf{80.4}             \\ \bottomrule
    \end{tabular}
\caption{BarbieGait improves upstream pose estimation, enabling better generalization to real-world dataset CCPG and SUSTech1K.}
\label{tab12}
\end{table*}

\section{Additional Ablation Studies}
\subsection{Effectiveness of Each Module}
Due to space limitations, Table 4 reports only the averaged performance on BarbieGait. A more comprehensive evaluation under all nine clothing conditions (THK1–THK9) is provided in Table~\ref{tab10}. As shown in the table, both GON-P3D and GON-FC contribute positively to cross-clothing robustness, each improving the baseline to varying degrees. The improvements are consistent across all settings and are particularly evident in the challenging THK7–THK9 conditions, highlighting the strong robustness and generalizability of our design.


\subsection{Ablations of the type of Normalization}
To further examine the effectiveness of our GON module, we conduct an additional ablation comparing it with commonly used normalization strategies, including Instance Normalization (IN), Batch Normalization (BN), and Layer Normalization (LN). By replacing GON with each standard normalization type while keeping all other components unchanged, we obtain a clear comparison in Table~\ref{tab11} that highlights the effects of \textit{Gait-Oriented Normalization} in challenging cloth-changing scenarios.

\section{Enhancing the Upstream Pose Estimator}

As an informative representation of human motion, 2D pose provides a stable and clothing-invariant description of gait through a fixed set of keypoints. From Table 3, the best-performing pose-based methods already surpass the best silhouette-based methods, indicating the strong potential of keypoint representations for cross-clothing gait recognition. However, current 2D pose estimation models~\cite{wang2020deep,xu2022vitpose,jiang2023rtmpose} are predominantly trained on action-oriented and motion-oriented datasets~\cite{ionescu2013human3,mahmood2019amass,andriluka2018posetrack,kanazawa2019learning,von2018recovering}. These datasets contain diverse activities and large motion amplitudes but involve only a limited number of subjects and lack clothing variation. In particular, they do not provide large-scale cross-clothing sequences for the same identity, making them insufficient as upstream supervision for cross-clothing gait recognition.

To enhance both the identity preservation capability and the generalization ability of upstream pose estimators under clothing variations, we extract one image every 10 frames from each sequence in BarbieGait, resulting in a 953K-image training set, which is significantly larger than the commonly used MS COCO~\cite{lin2014microsoft} dataset with 150K images. Beyond its scale, BarbieGait also provides richer gait-specific motion patterns and identity-consistent clothing variations, aligning more closely with the requirements of downstream gait tasks.

For a fair comparison with MS COCO–based methods, we ensure that our pose estimator predicts the same number of keypoints and maintains comparable body semantics. As shown in Figure~\ref{fig5}, both the COCO-17 format (the standard 17-keypoint layout used in MS COCO) and our Barbie-17 format (the 17-keypoint layout defined in BarbieGait) contain the same total number of joints. Following the keypoint selection strategy adopted in~\cite{lang2025beyond}, we keep the COCO-style semantics for the 12 skeletal joints of the limbs and torso, while replacing the original COCO head keypoints with 5 mesh-derived head landmarks that provide more stable and anatomically reliable supervision.

For training, we adopt ViTPose-H~\cite{xu2022vitpose} implemented in MMPose~\cite{mmpose2020} as our backbone. The model is initialized with MAE~\cite{he2022masked} pre-trained weights and trained with default MMPose settings: an input size of $256 \times 192$, the AdamW~\cite{reddi2019convergence} optimizer with a learning rate of 1e-3, UDP~\cite{huang2020devil} post-processing, a batch size of 512, and 20 training epochs with learning rate decay at epochs 8 and 16.

Ultimately, by deploying our retrained pose estimation model, we achieve substantial performance improvements on publicly available RGB gait datasets, including CCPG and SUSTech1K. By incorporating our retrained pose model into the SkeletonGait~\cite{fan2024skeletongait} pipeline, we obtain new state-of-the-art performance on these real-world benchmarks, as shown in Table~\ref{tab12}.

\section{Additional Visualization}
\subsection{Visualization of Synthetic Images }
\begin{figure*}[!htbp]
    \centering
    \includegraphics[width=0.95\textwidth]{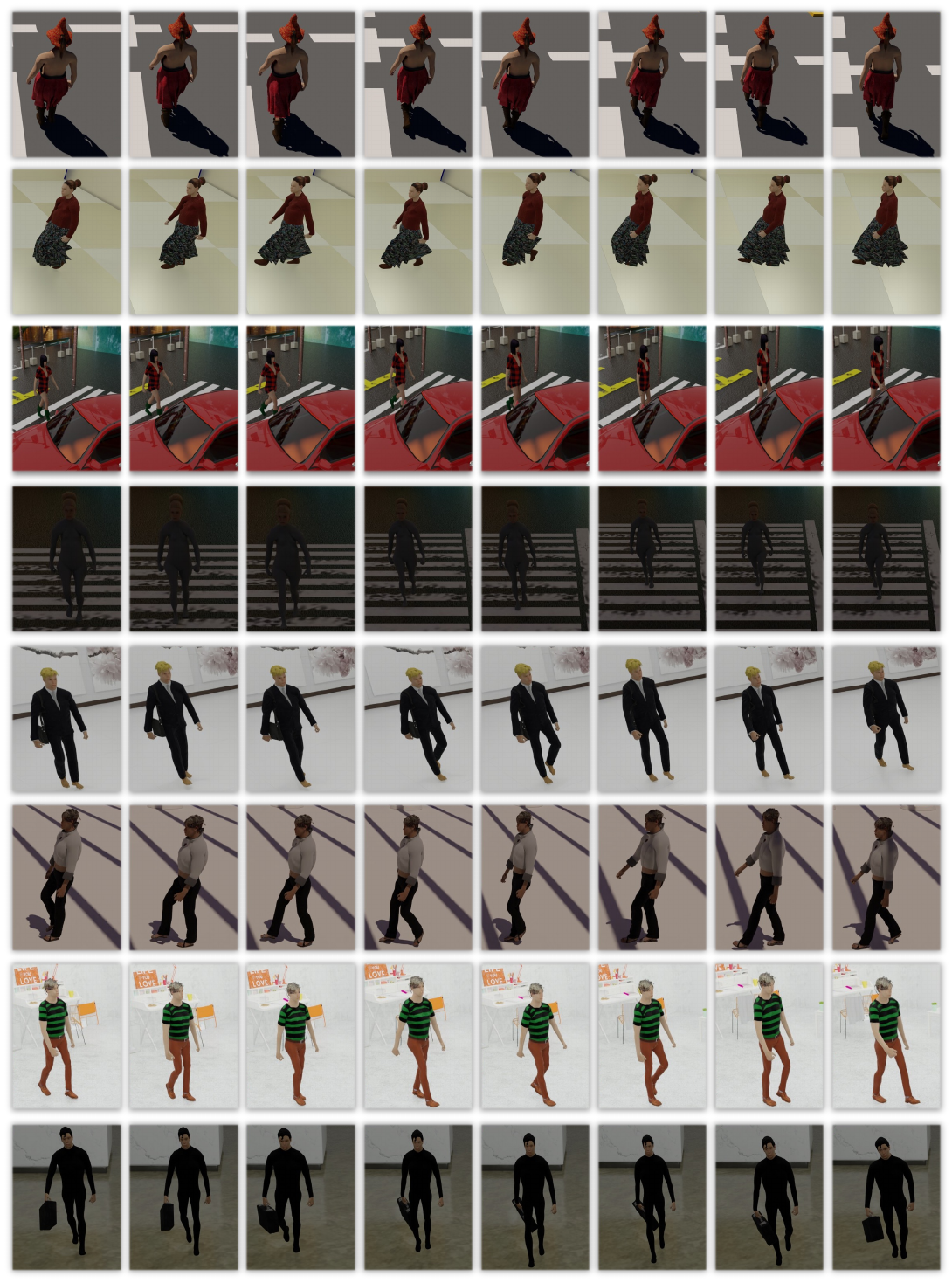}
    \caption{The illustration of our synthesized images. Our synthetic images are rendered in different scenes, realistic lighting conditions, diverse clothing conditions, and natural occlusions.}  
    \label{fig8}  
\end{figure*}

We also present additional synthetic images from BarbieGait in Figure \ref{fig8}. We select two subjects, each shown in four different scenes, including indoor scenes and outdoor scenes under varying lighting conditions. We simulate \textbf{lighting variations} as realistically as possible by setting appropriate lighting conditions in both indoor and outdoor environments. For example, indoors, we simulated incandescent lighting conditions, while outdoors, we mimicked sunlight variations. In addition, the subjects naturally interact with scene objects, resulting in realistic \textbf{occlusions}. The realistic scene and lighting simulation, combined with real gait data sources, ensure the validity and authenticity of BarbieGait as a synthetic gait dataset.

\subsection{Visualization of Heatmaps}
\begin{figure}[t] 
    \centering
    \includegraphics[width=1.0\columnwidth]{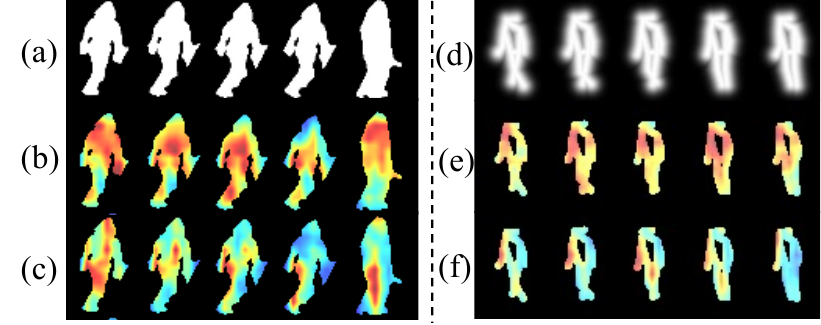}  
    \caption{Visualization of heatmaps in Silhouette-based (a)-(c) and Pose-based methods (d)-(f). (b) and (e) show activation heatmaps of DeepGaitV2 and SkeletonGait overlaid on the silhouette. (c) and (f) show the effect with GaitCLIF.}  
    \label{fig7}  
\end{figure}

Figure \ref{fig7} provides qualitative insights into the impact of clothing. As clothing complexity increases, crucial gait information is obscured while irrelevant clothing details are introduced in silhouette-based methods. If the model focuses excessively on clothing, it fails to capture critical gait features. Comparing Figure \ref{fig7}(b) and \ref{fig7}(c), DeepGaitV2 focuses more on the clothing areas, while GaitCLIF concentrates on unoccluded and discriminative areas such as body joints and edges. This demonstrates that GaitCLIF effectively removes clothing stylization and guides the model to focus more on gait information that is independent of clothing. Pose-based methods are less affected by clothing than silhouette-based methods but provide limited useful information. This requires the model to focus more on fine-grained features. As shown in (e), SkeletonGait mainly activates the entire skeleton map, while with the help of GaitCLIF, (f) shows a shift toward dynamic joint regions, learning more discriminative gait features.

\subsection{Visualization of Diverse Clothing }
In this section, we provide additional visualizations of the diverse clothing used in BarbieGait, as shown in Figure~\ref{fig6}. The wardrobe includes a wide range of apparel and appearance attributes, covering various hairstyles, tops, pants, skirts, shoes, and carried objects. Each category contains roughly 100 individual items, enabling more than 200,000 theoretically valid clothing combinations.

During outfit generation, we follow common real-world dressing conventions and further apply manual filtering to remove implausible combinations as well as cases exhibiting mesh–cloth penetration (i.e., garments intersecting with the body surface). These measures ensure that the clothing variations in BarbieGait are both realistic and suitable for gait recognition studies under cloth-changing conditions.



\end{document}